\documentclass[conference]{IEEEtran}
\IEEEoverridecommandlockouts
\usepackage{cite}
\usepackage{amsmath,amssymb,amsfonts}
\usepackage{graphicx}
\usepackage{textcomp}
\usepackage{xcolor}
\usepackage{threeparttable}
\usepackage{algorithm}
\usepackage{algorithmicx}
\usepackage{algpseudocode}

\usepackage{multirow} 

\def\BibTeX{{\rm B\kern-.05em{\sc i\kern-.025em b}\kern-.08em
    T\kern-.1667em\lower.7ex\hbox{E}\kern-.125emX}}
\begin{document}

\title{Order-sensitive Neural Constituency Parsing}
\author{\IEEEauthorblockN{Zhicheng Wang, Tianyu Shi, Liyin Xiao, Cong Liu\textsuperscript{*}\thanks{Cong Liu is the corresponding author.}}
\IEEEauthorblockA{\textit{School of Computer Science and Engineering} \\
\textit{Sun Yat-sen University}\\
Guangzhou, China \\
\{wangzhch23, shity3, xiaoly28\}@mail2.sysu.edu.cn, liucong3@mail.sysu.edu.cn}
}

\maketitle

\begin{abstract}
We propose a novel algorithm that improves on the previous neural span-based CKY decoder for constituency parsing. In contrast to the traditional span-based decoding, where spans are combined only based on the sum of their scores, we introduce an order-sensitive strategy, where the span combination scores are more carefully derived from an order-sensitive basis. Our decoder can be regarded as a generalization over existing span-based decoder in determining a finer-grain scoring scheme for the combination of lower-level spans into higher-level spans, where we emphasize on the order of the lower-level spans and use order-sensitive span scores as well as order-sensitive combination grammar rule scores to enhance prediction accuracy. We implement the proposed decoding strategy harnessing GPU parallelism and achieve a decoding speed on par with state-of-the-art span-based parsers. Using the previous state-of-the-art model without additional data as our baseline, we outperform it and improve the F1 score on the Penn Treebank Dataset by 0.26\% and on the Chinese Treebank Dataset by 0.35\%. 
\end{abstract}


\section{Introduction}

The application of the neural span-based decoder \cite{MitchellStern2017AMS} significantly increases the constituency parsing accuracy. The majority of span-based constituency parsers used today are based on the encoder-decoder architectures \cite{kitaev2018}, in which the encoder converts input sentences to vector arrays representing words and then uses a scorer to calculate a score for each span, and finally, the decoder uses the span scores to construct a constituency tree. After replacing the original LSTM model with a transformer-based  \cite{kitaev-etal-2019-multilingual} encoder, the accuracy of constituency parsing is increased further.

The introduction of neural networks simplifies the decoder of the algorithm. While the traditional CKY algorithm matches each non-terminal node in the parse tree with the grammar rule formed by the left child node, right child node, and their parent node, neural CKY utilizes a span scorer to perform chart-parsing, which applies a dynamic programming algorithm to determine the maximum value of the subtree score from bottom to top in order to construct the globally optimal parse trees. This is an efficient approach that results in high parsing performance.

\begin{figure*}[htbp]
\centerline{\includegraphics[width=0.85\textwidth]{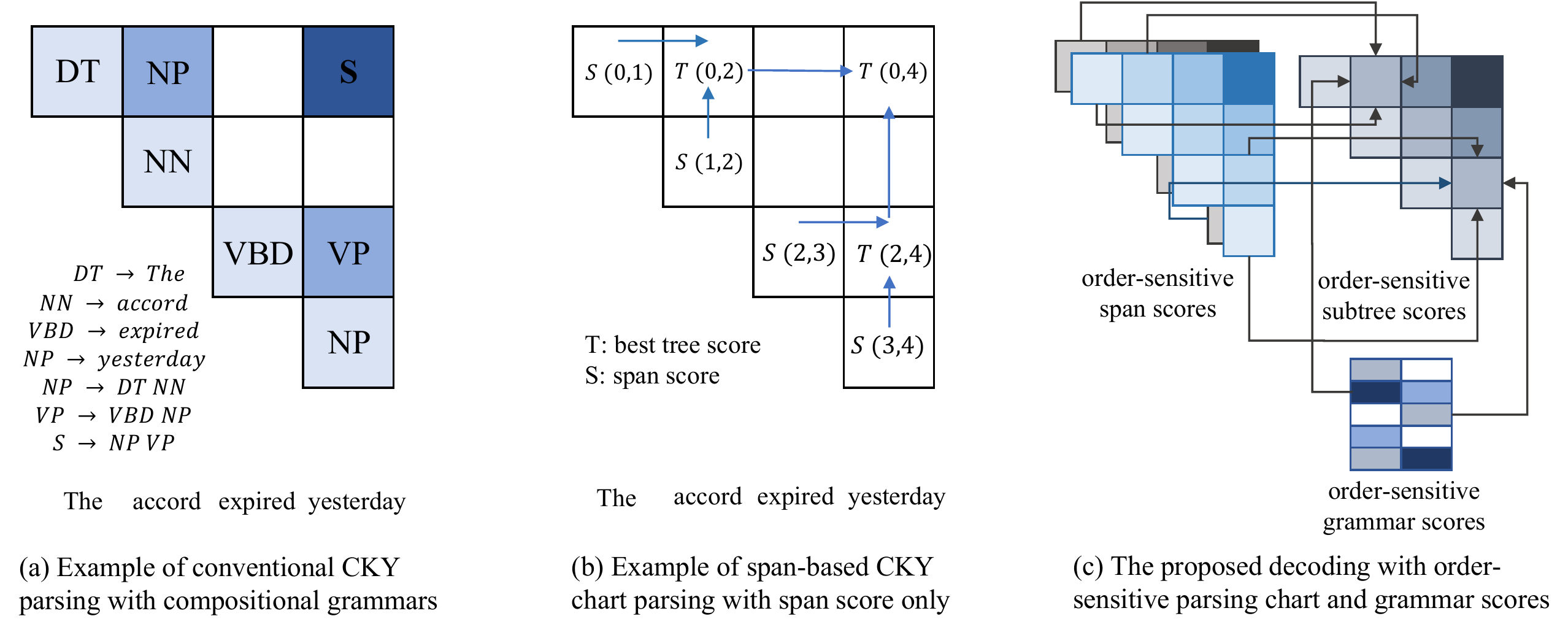}}
\caption{Comparisons of Classic, Span-based, and Order-sensitive CKY}
\label{fig1}
\end{figure*}

However, this span-based neural decoder relies on the neural scorer to ensure the correctness of the compositional spans and the order between the left and right child spans in these compositions, which is not explicitly modeled in the neural network. As illustrated in Fig. \ref{fig1}, the neural scorer will give a child span the same score without considering its order; therefore, correctness is not guaranteed. But the traditional CKY algorithm elicits spans based on grammar, which includes information about the label and the order of its child spans. For example, the grammar rule NP $\rightarrow$ DT NN implies that for parent node NP, DT is the left node, and NN is the right node. Because there is no rule for DT as a right child, the traditional method implicitly contains order information.

To attack this problem, we propose an order-sensitive scorer in our decoder. In our model, we carefully consider how the order among the children can affect the score of a subtree. Specifically, we determine the per root label score of each subtree based on four values: the per-label span score of the left split, the per-label span score of the right split, the per-label score of the parent span, and the score of composition grammar rule which consists of the labels of the left and right child spans and that of the parent span. To model the difference between the order of spans, we define two scores for each of the scores above.

After incorporating this strategy, as depicted in Fig. \ref{fig1}, the tree score of the word \textit{The} as the left node should be greater than the tree score of it as the right node, so that the order information is factored in when constructing the parse tree.

Our algorithm can be computed in parallel easily. We modify the span-based neural constituency parsing model \cite{MitchellStern2017AMS} and use the previous state-of-the-art model which is proposed by \cite{KhalilMrini2020RethinkingST} as the baseline for our experiment. Our model's F1 score is 0.26\% higher than the baseline on the PTB \cite{MitchellMarcus1993BuildingAL} dataset and 0.35\% higher on the CTB \cite{NaiwenXue2005ThePC} dataset.

\section{Related Work}
\subsection{CKY Algorithm}

The CKY algorithm \cite{JohnCocke1969ProgrammingLA,TadaoKasami1965AnER,DanielHYounger1967RecognitionAP} is the first polynomial-time parsing algorithm applicable to ambiguous CFGs that permit multiple derivations. The conventional CKY parsing algorithm can enumerate all possible parse trees for a given sentence, but it usually does not properly disambiguate between them. A neural application of the CKY algorithm (chart parsing) is proposed to handle the disambiguation. The idea behind neural CKY parsing algorithms \cite{MitchellStern2017AMS} is to use a modified version of CKY to combine constituent scores and find the optimal parse tree. Our approach improves the chart parsing algorithm and introduces the order-sensitive scores which enhance the information about the span order.

\subsection{Transition-Based Parsing}
Transition-based statistical parsing assigns a score to each shift or reduce actions made during the parsing process, and then chooses the parsing tree constructed from the highest scoring sequence of decisions. It is challenging for transition-based parsing to handle natural languages that are difficult to parse due to their complex semantics and long dependencies. Early attempts use classifier \cite{sagae-lavie-2005-classifier} and best-first search \cite{sagae-lavie-2006-best}, while recent work utilizes specific tags \cite{kitaev-klein-2020-tetra} and dynamic oracles \cite{JamesCross2016SpanBasedCP} to improve accuracy.

\subsection{Span-Based Neural Constituency Parsing}

Several constituency parsers proposed in the literature in recent years can be categorized as span-based neural constituency parsers. They train neural networks to assign a score to each span and then use a modified version of CKY to find an optimal parse tree, which contains spans whose scores sum up to a maximum value. In \cite{kitaev2018}, the parser combines a chart decoder with a sentence encoder based on self-attention. Label attention layer \cite{KhalilMrini2020RethinkingST} improves sentence embeddings by appending extra embeddings to the sentences that provide task-specific information in a way similar to prompting \cite{DBLP:journals/corr/abs-2107-13586}. Head-driven phrase structure grammar \cite{IvanASag1994HeaddrivenPS} benefits from a uniform formalization that allows for the representation of rich contextual syntactic and semantic meanings. Recursive semi-Markov model \cite{XinXin2021NaryCT} utilizes the 1-order semi-Markov model to predict the immediate children sequence of a constituent candidate. Our proposed framework can be computed in parallel by GPU without introducing higher complexity.

\section{Methodology}

\subsection{Order-sensitive Decoding Algorithm}

We count the number of labels after splitting to left child and right child after binarization following the same method in \cite{kitaev2018} for the training set and the test set of PTB, as shown in Table \ref{table1}, which demonstrates a significant difference in the occurrence probability of label's order.

\begin{table}[htbp]
\centering
\caption{Numbers of labels as left and right children in PTB dataset}
\begin{threeparttable}
\begin{tabular}{|l|l|l|l|l|}
\hline \multirow{2}{*}{\bf Label\tnote{a}} &\multicolumn{2}{|c|}{ \bf Train} & \multicolumn{2}{|c|}{\bf Test}\\
\cline{2-5}
 &\bf L &\bf R &\bf L &\bf R\\ \hline
NP     & 82466 & 114615 & 5016 & 6692 \\
VP     & 253   & \textbf{68520}  & 16   & \textbf{4173} \\
PP     & 333   & \textbf{51629}  & 21   & \textbf{3042} \\
S      & 169   & \textbf{31025}  & 4    & \textbf{1874} \\
SBAR   & 46    & \textbf{15443}  & 5    & \textbf{952}  \\
WHNP   & \textbf{6777}  & 394    & \textbf{395}  & 25   \\
ADJP   & 1408  & 5175   & 89   & 348  \\
QP     & \textbf{2442}  & 185    & \textbf{136}  & 5    \\
WHADVP & \textbf{1934}  & 0      & \textbf{124}  & 0    \\
ADVP   & 1103  & 2185   & 64   & 148  \\
\hline
\end{tabular}
\begin{tablenotes}    
    \footnotesize               
    \item[a] The ten most frequently occurring labels are selected in the table.
\end{tablenotes}   
\end{threeparttable}
\label{table1}
\end{table}

Table \ref{table1} show that the occurrence probability between left and right child nodes is similar for NP, ADVP in PTB. However, there are more labels with a greater difference of probability for its order, such as VP, PP, SBAR, whose number of occurrences as right child node is at least 10 times that of left, whereas the number of occurrences as left child node for QP, WHNP, WHADVP is much greater than that of the right child node. In this case, it is illogical to assign the same score to each span label when it represents in a different order. For example, WHADVP, whose span score as a right child node should be as lower as possible. As a result, we develop an order-sensitive algorithm to address this issue.

After the encoder section, we obtain the score for each span from the encoder. The purpose of the Order-sensitive Decoding Algorithm is to build a parse tree from the scores. The parse tree’s score is calculated by summing the subtree scores of the constituent tree:
\begin{equation}
    t(T)=\sum_{(i,j,l)\in T} t(i,j,l)
\end{equation}
where $T$ represents the parsing tree, and $t$ is the subtree score.

The above equation can be used to represent any type of parse tree obtained through any partition, and the most reasonable parse tree is the final parse tree with the highest tree score:

\begin{equation}
    T^*=\arg\max\limits_T t(T)
\end{equation}

When span size is 1,

\begin{equation}
    t_{best}(i,j)=\max\limits_l s(i,j,l)
\end{equation}

When span size is larger than 1,

\begin{equation}
\begin{aligned}
t_{best}(i,j)&=\max\limits_l s(i,j,l) \\
&+ \max\limits_k[t_{best}(i,k) + t_{best}(k,j)]
\end{aligned}
\end{equation}
where $l$ is the label, $s(i,j,l)$ is the span score of span $(i,j)$. 

However, the preceding strategy disregards the difference between the order of labels and assigns identical scores to each span as the left or right subtree.

To address this problem, we use neural networks to compute distinct scores $s(i,j,l,o)$ for the same span regarding their order where $o$ is the order of a span. In a binarized tree, $o$ is $L$ for the left child span and $R$ for the right child span.

Secondly, we elicit the compositional grammar from the training set, based on which we create a list of grammar rule scores $g(G,o)$, where $o$ represents the order of the parent node in the grammar rule. The grammar rule score chart is randomly initialized and is the parameter in the parsing model.

With the order sensitive span scores and the grammar rule scores introduced above, we can define our order-sensitive decoding as follows:

When span size is 1:

\begin{equation}
    t(i,j,l,o)=s(i,j,l,o)
\end{equation}

When span size is larger than 1, 

\begin{equation}
\begin{aligned}
t(i,j,l,o)&=s(i,j,l,o)+\max\limits_{i<k<j,\ G:\ l\,\rightarrow\,l_1\;l_2}\\&[t(i,k,l_1,\!L)\!+\!t(k,j,l_2,\!R)\!+\!g(G,o)]
\end{aligned}
\end{equation}

The optimal tree is given by:

\begin{equation}
    t_{best}(0,n) = \max\limits_{l} t(0,n,l,L)
\end{equation}
since the order as a child is meaningless for the root node.

Our order-sensitive decoding algorithm for binarized tree constituency parsing is shown in the Algorithm \ref{A1}. In the algorithm, each subtree is the sum of four scores. Compared with the conventional chart parsing, our span scores $s(i,j,l,o)$ are order-sensitive. For binarized trees, they are given by $s(i,j,l,L)$ and $s(i,j,l,R)$ for left spans and right spans respectively. For the subtree over the left child span $(i,k)$, we use the order sensitive score $t(i,k,l_1,L)$, while for the subtree over the right child span, we use $t(k,j,l_2,R)$. Finally, we have a grammar rule score $g(G,o)$ for each grammar rule $G:l\rightarrow l_1 l_2$, which is also order-sensitive.

\begin{algorithm}
\caption{Order-sensitive Decoding Algorithm for Binarized Trees}
\begin{algorithmic}[1]
\Require{Sentence length $n$, ordered span scores $s(i,j,l,o), 0 \le i < j\le n$, order rule scores $g(G,o)$, for grammar rule $G:\;l\rightarrow l_1\, l_2$.}
\Ensure{Tree $T^*$ with maximum value $t(T^*)$.}
\For{$w \leftarrow 1$ \rm \textbf{to} $n$} \Comment{span size}
    \For{$i \leftarrow 0$ \rm \textbf{to} $n-w+1$}
        \State $j\leftarrow i+w$
        \For{\textbf{all labels} \rm $l$}
            \If{$w=1$} \Comment{span size 1}
                \State $t(i,j,l,o) \leftarrow s(i,j,l,o)$ \Comment{$o$ in $L,R$}
            \Else \Comment{span size larger than 1}
                \State $t(i,j,l,o) \leftarrow s(i,j,l,o)+\max\limits_{i<k<j,\ G:\,l\,\rightarrow\,l_1\;l_2}[\,t(i,k,l_1,L)+t(k,j,l_2,R)+g(G,o)\,]$
            \EndIf
        \EndFor
    \EndFor
\EndFor
\State $t_{best}(0,n) \leftarrow \max\limits_{l} t(0,n,l,L)$ \Comment{$t(T^*)$}
\State Trace back the tree $T^*$
\end{algorithmic}
\label{A1}
\end{algorithm}

In our implementation, all computations within the inner loop of the same span are implemented using batchified GPU operations. The key concept is to compress the computation of the same span size into a large tensor calculation. Because all parameters involved in the computation for subtree scores with the same span size are independent of each other, the calculation of one sentence can be accomplished in $n$ steps, where $n$ is the length of the sentence.

\subsection{Model}

\begin{figure*}[t]
\centering
\includegraphics[width=0.7\textwidth]{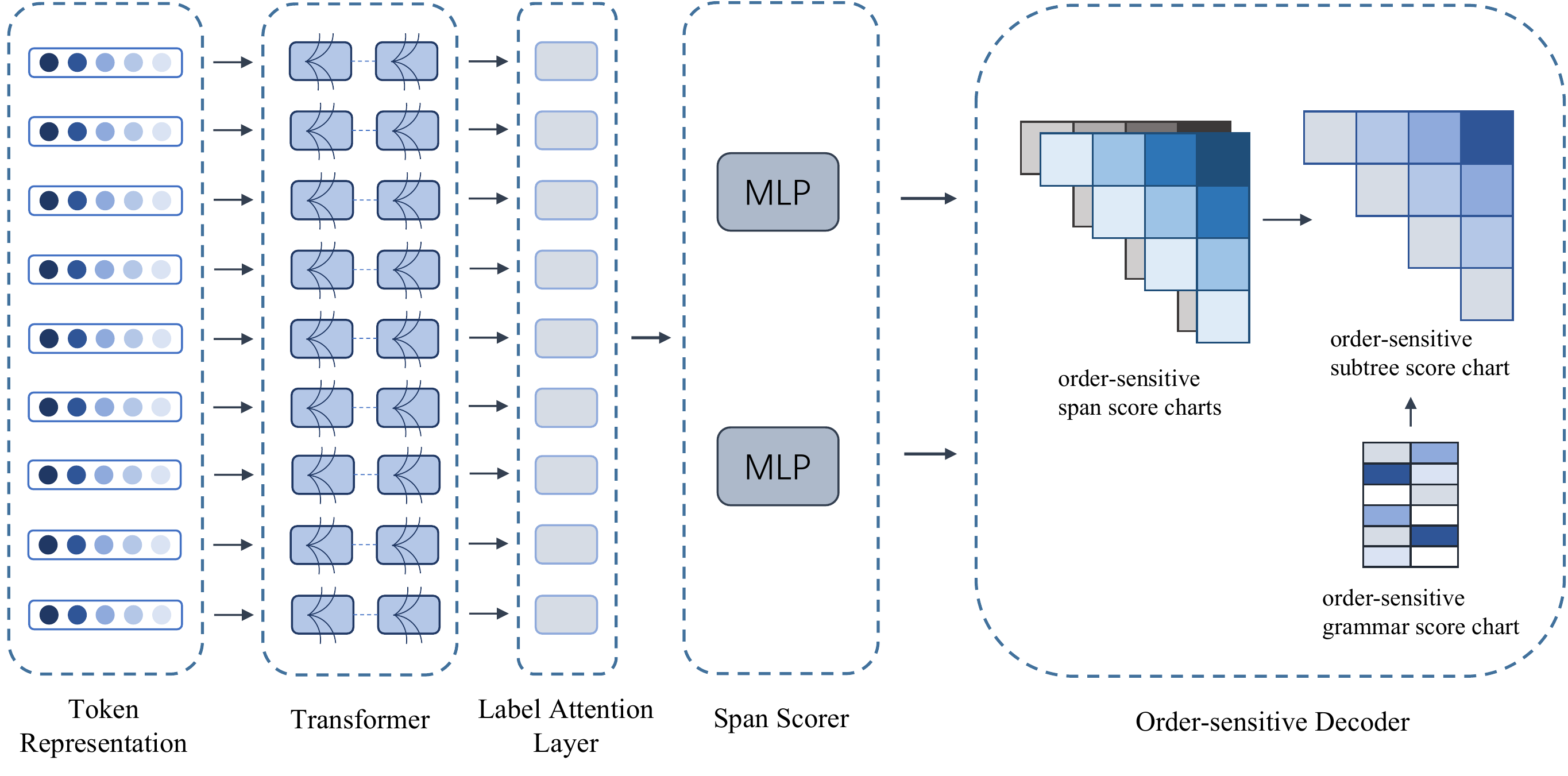}
\caption{Structure of Our Model}\label{fig2}
\end{figure*}

Our parser, which is shown in the Fig. \ref{fig2}, is based on the encoder-decoder model \cite{kitaev2018}, with some improvements resulting from subsequent research. Our model uses XLNet \cite{ZhilinYang2019XLNetGA}, incorporates a label attention layer \cite{KhalilMrini2020RethinkingST} to improve self-attention \cite{AshishVaswani2017AttentionIA}, and augments the scoring layer and decoder \cite{MitchellStern2017AMS} with our order-sensitive decoder.

The model passes through three modules in the encoder: Token Representation shown in Section \ref{sec:TR}, which converts the word to a vector; Attention Layers shown in Section \ref{sec:AL}, which augments the vector with content and position information; and Scoring Layer shown in Section \ref{sec:SL}, which calculates the vector corresponding to the span and scores it.

Following the encoder section, we obtain the score for each span from the scorers. We construct the parse tree from the span score using our order-sensitive decoding algorithm in the decoder shown in Section \ref{sec:decoder}.

\subsubsection{Token Representation\label{sec:TR}}
Our model's token representation is bipartite. The initial step is to vectorize the words in the sentence $s$ using pretrained language models. 

\begin{equation}
    s=\{w_1,w_2,w_3,...w_n\}
\end{equation}

This section makes use of two pre-trained language models: XLNet \cite{ZhilinYang2019XLNetGA} for the Penn Treebank Dataset \cite{MitchellMarcus1993BuildingAL} and BERT-Chinese \cite{10.1109/TASLP.2021.3124365} for the Chinese Treebank Dataset \cite{NaiwenXue2005ThePC}. This step extracts the content of each word using pre-trained language models.

\subsubsection{Attention Layers\label{sec:AL}}

Following the initial embedding of sentences, it enters the attention layers, which are divided into three sections. First, the word vector enters the multilevel embedding module, where the sentences' content and position features are bound to the vector:

\begin{equation}
    E_n = [ v_n ; p_n ]
\end{equation}
where $v_n$ means the token representation passed from the previous section, $p_n$ is the position embedding.

The processed vector is then passed through 12 layers of the normal self-attention layer, which is initialized following the method from \cite{AshishVaswani2017AttentionIA}. After that, the processed vector is given to a label attention layer \cite{KhalilMrini2020RethinkingST}. This layer of attention assigns a query vector to each semantic label, thereby substituting the label for the attention heads.

\subsubsection{Scoring layer\label{sec:SL}}
In the scoring layer, span representations adhere to the definitions from \cite{kitaev2018} and \cite{DavidGaddy2018WhatsGO}. The span vector $v(i,j)$ is calculated based on the span containing the words from $i$ to $j$:

\begin{equation}
    v(i,j) = [\mathop{h_j}\limits ^{\rightarrow}-\mathop{h_i}\limits ^{\rightarrow};\mathop{h_{j+1}}\limits ^{\leftarrow}-\mathop{h_{i+1}}\limits ^{\leftarrow}]
\end{equation}
where $\mathop{h_i}\limits ^{\rightarrow}$ and $\mathop{h_i}\limits ^{\leftarrow}$ is the backward and forward expression of the i-th word embedding $E_n$. The backward and forward representation can be obtained by splitting the representation in half. The span vector $v(i,j)$ in the previous work is then passed through a multilayer perceptron \cite{rosenblatt1961principles} span classifier with two fully connected layers and a single ReLU activation function \cite{pmlr-v15-glorot11a}, the output dimensionality of which is equal to the number of possible non-terminal labels:

\begin{equation}
    s(i,j,\cdot)=W_2 ReLU(LN(W_1 v(i,j))
\end{equation}
where $W_1$ and $W_2$ are the fully connected layers, $s(i,j)$ is the score of span for the specific label, $LN$ is the layer normalization. Because the order needs to be considered in this section, each label of each span will have two scores, one for the left subtree and one for the right subtree, resulting in a different scoring equation:

\begin{equation}
    s(i,j,o,\cdot)\!=\!W_{o,2} ReLU(LN(W_{o,1} v(i,j))
\end{equation}
where $o$ is the order of spans, $W_{o,2}$ and $W_{o,1}$ are two fully connected layers for the specific order.

In our model, we use two MLPs to generate span score of different order separately.

\subsubsection{\label{sec:decoder}Decoder}

In this section, we use our improved order-sensitive neural CKY algorithm to form the decoder, which is parallelized in the computation. The model converts the input score into a parse tree and extracts the labels of tree nodes. We use the ordered span score chart from the span scorer and the order rule score generated by the grammar rules. The rule score in the rule score chart, in particular, can be learned during the training process.

\subsubsection{Loss}

The model is trained to predict the correct parse tree $T^*$. In the model, we can predict different tree $T$ with different scores. It can be inferred that the following constraint must be satisfied: 

\begin{equation}
    t(T^*)\ge t(T)+\Delta(T,T^*)
\end{equation}
where $\Delta$ is the Hamming loss for the label of spans, $t(T^*)$ is the score of the correct tree. The loss function in the model is the hinge loss:

\begin{equation}
    L = \max(\max_T(t(T)+\Delta(T,T^*)-t(T^*),0)
\end{equation}

In the training step, we minimize the loss to achieve the optimal model.

\section{Experiment}

We analyze our model on the English Penn Tree-bank (PTB) \cite{MitchellMarcus1993BuildingAL} and Chinese Treebank 5.1 (CTB) \cite{NaiwenXue2005ThePC} which has the data splitting from \cite{liu-zhang-2017-shift} and left-branching binarization. Following normal procedure, we apply the EVALB \cite{sekine1997evalb} method for assessing the F1 score.

\subsection{Setup}
In the English experiment (PTB), we use large-cased pretrained XLNet, followed by a 12 self-attention layer. For the label attention layer, we follow the setting of \cite{KhalilMrini2020RethinkingST}, using 112 heads (one per syntactic category), and for the MLP, we utilize two two-layer MLPs with the hidden layer of dimension 250.

In our Chinese experiments (CTB), we use BERT as our pre-training model. Except for the label attention layer, which uses 64 heads \cite{KhalilMrini2020RethinkingST}, the rest of parameters are consistent with the English experiments.

Our specific parameters are depicted in Table \ref{para}.

\begin{table}[htbp]
\begin{center}
\caption{Hyper-parameters in our experiment.}
\begin{threeparttable}
\begin{tabular}{|l|l|}
\hline \bf Parameter & \bf value\\ \hline
batch size & 32\\
decay factor & 0.5\\
max decay & 3\\
attention layer & 12\\
LAL head & 112 / 64\tnote{a}\\
learning rate & 3e-5\\
decay patience & 5\\
dropout & 0.2\\
MLP layer & 2\\
MLP hidden layer & 250\\
rule chart min & -1e6\\
\hline
\end{tabular}
\begin{tablenotes}    
    \footnotesize               
    \item[a] 112 heads for PTB and 64 heads for CTB.
\end{tablenotes}   
\end{threeparttable}
\label{para}
\end{center}

\end{table}

\subsection{Performance}

Table \ref{ptb} indicates the results of our proposed parser on the PTB dataset. The baseline model in the table refers to the result after removing the order-sensitive component, that is, using the XLNet and the label attention layer in the encoder, and using ordinary neural CKY parsing in the decoder. Our model has 0.26\% improvements on PTB compared with the baseline. In comparison to other models that do not incorporate data from dependency parsing, our model's accuracy is comparable to the state-of-the-art model.

\begin{table}[htbp]
\begin{center}
\caption{Constituency Parsing Performance on PTB test set.}
\begin{threeparttable}
\begin{tabular}{|l|c|c|c|}
\hline \bf Model & \bf Precision &\bf Recall &\bf F1 \\ \hline
Syntactic Distance \cite{shen-etal-2018-straight}& 92.00 & 91.70 & 91.80 \\
Local Models \cite{teng-zhang-2018-two}& 92.50 & 92.20 & 92.40 \\
Sequence Tagging \cite{vilares-etal-2019-better}& - & - & 90.60 \\
HPSG \cite{zhou-zhao-2019-head}\tnote{a}& 96.21 & 96.46 & 96.33 \\
Multilingual\cite{kitaev-etal-2019-multilingual}& 95.73 & 95.46 & 95.59 \\
CRF\cite{ijcai2020-0560}& 95.85 & 95.53 & 95.69 \\
LAL\cite{KhalilMrini2020RethinkingST}\tnote{a}& 96.53 & 96.24 & 96.38 \\
Linearization\cite{wei-etal-2020-span}& 95.50 & \textbf{96.10} & 95.80  \\
Recursive semi-Markov\cite{XinXin2021NaryCT}& \textbf{96.29} & 95.55 & \textbf{95.92} \\
\hline
Baseline & 95.83 & 95.42 & 95.61 \\
Our model & 95.97 & 95.77 & 95.87 \\
\hline
\end{tabular}
\begin{tablenotes}    
    \footnotesize               
    \item[a] Use the data from Dependency Parsing.
\end{tablenotes} 
\end{threeparttable}
\label{ptb} 
\end{center}

\end{table}

Table \ref{ctb} indicates the results of CTB dataset. The correctness of our model is 0.35\% higher than the baseline. Among models without using dependency parsing data, our model establishes new state-of-the-art results.

\begin{table}[htbp]
\begin{center}
\caption{Constituency Parsing Performance on CTB 5.1 test set.}
\begin{threeparttable}
\begin{tabular}{|l|c|c|c|}
\hline \bf Model & 
\bf Precision &\bf Recall &\bf F1 \\ \hline
Syntactic Distance \cite{shen-etal-2018-straight}& 86.60 & 86.40 & 86.50 \\
Local Models\cite{teng-zhang-2018-two}& 87.50 & 87.10 & 87.30 \\
Sequence Tagging\cite{vilares-etal-2019-better}& - & - & 85.61 \\
HPSG\cite{zhou-zhao-2019-head}\tnote{a}& 92.33 & 92.03 & 92.18 \\
Multilingual\cite{kitaev-etal-2019-multilingual}& 91.96 & 91.55 & 91.75 \\
CRF\cite{ijcai2020-0560}& 92.51 & 92.04 & 92.27 \\
LAL\cite{KhalilMrini2020RethinkingST}\tnote{a}& 93.45 & 91.85 & 92.64 \\
Linearization\cite{wei-etal-2020-span}& 92.70 & 92.20 & 92.40 \\
Recursive semi-Markov \cite{XinXin2021NaryCT}& \textbf{92.94} & 92.06 & 92.50 \\
\hline
Baseline & 91.99 & 92.33 & 92.16 \\
Our model & 92.55 & \textbf{92.46} & \textbf{92.51} \\
\hline
\end{tabular}
\begin{tablenotes}    
    \footnotesize               
    \item[a] Use the data from Dependency Parsing.
\end{tablenotes} 
\end{threeparttable}
\label{ctb}
\end{center}
\end{table}

\subsection{Ablation Study}

\begin{table}[htbp]
\begin{center}
\caption{Ablation Study on PTB test set.}
\begin{threeparttable}
\begin{tabular}{|l|c|c|c|}
\hline \bf Model & 

\bf Precision &\bf Recall &\bf F1 \\ \hline
Baseline & 95.83 & 95.42 & 95.61 \\
Ordered\;span\;score& 95.92 & 95.65 & 95.79 \\
Order\,\&\,Grammar rule\,score& \textbf{95.97} & \textbf{95.77} & \textbf{95.87} \\
\hline
\end{tabular}
\end{threeparttable}
\label{ablation}
\end{center}

\end{table}

We next conduct an ablation study to discover more about the contributions of various components in our proposed model.

\paragraph{Impact of grammar rule scores.}
In our parsing model, we use the ordered span scores and the grammar rule scores to calculate the parsing tree, so we remove the grammar rule scores to evaluate its impact. That is, we simplify our decoder using the following subtree score:

\begin{equation}
\begin{aligned}
t(i,j,o)&=\max\limits_{l}[s(i,j,l,o)\\
&+ \max\limits_{i<k<j}(t(i,k,L)+t(k,j,R))]\;
\end{aligned}
\end{equation}

The results depicted in the Table \ref{ablation} clearly show that the introduction of the grammar rule chart improves the F1 score of the constituency parsing.

\paragraph{Impact of order-sensitive decoding.}
We can evaluate the impact of the order-sensitive decoding by comparing the results of the baseline with the simplified model with ordered span score. The simplified model can have 0.18\% improvement of F1 score.

\subsection{Speed Comparison}

We average the results of parsing speed from 20 trials which are measured on the PTB test set. With a single Tesla V100, the average processing rate is 65 sentences per second. Table \ref{speed} shows the speed comparison between our model and the baseline. The speed is slightly slower in live testing, but considering the increase in accuracy, the processing speed is acceptable.

\begin{table}[htbp]
\begin{center}
\caption{Speed comparison on PTB test set.}
\begin{threeparttable}
\begin{tabular}{|l|c|}
\hline \bf Model & \textbf{Sents/sec} \\ \hline
Baseline &  103 \\
Ordered span score & 89 \\
Ordered \& grammar rule score&  65\\
\hline
\end{tabular}
\end{threeparttable}
\label{speed}
\end{center}
\end{table}

\section{Conclusion}

In this paper, we introduce the order-sensitive decoding algorithm to the neural CKY constituency parser which combines order-sensitive span scores and ordered grammar rule scores, with the benefit of enhancing the prediction accuracy for the parse trees. We discuss the advantages and necessity of the order-sensitive decoding in the label statistics and ablation study sections. Our model is not time-consuming compared with previous model computing on GPU. Experiments demonstrate that our model outperforms the baseline model, which incorporates elements of the previous state-of-the-art model on the PTB and CTB datasets.

\bibliographystyle{IEEEtran}
\bibliography{main.bib}

\end{document}